\documentclass[lettersize,journal]{IEEEtran}
\usepackage{amsmath,amsfonts}
\usepackage{algorithmic}
\usepackage{algorithm}
\usepackage{array, multirow}
\usepackage[caption=false,font=normalsize,labelfont=sf,textfont=sf]{subfig}
\usepackage{textcomp}
\usepackage{stfloats}
\usepackage{url}
\usepackage{balance}
\usepackage{verbatim}
\usepackage{graphicx}
\usepackage{cite}
\usepackage[colorlinks=true, allcolors=blue]{hyperref}

\usepackage{orcidlink}

\usepackage{color,soul}

\newcommand{\highlighting}[1]{%
    {%
    \sethlcolor{white}%
    \hl{#1}%
    }%
}
\begin{document}

\title{PolSAR Image Classification using a Hybrid Complex-Valued Network (HybridCVNet)}

\author{Mohammed~Q.~Alkhatib$^{\orcidlink{0000-0003-4812-614X}}$, \textit{Senior Member}, IEEE}

\maketitle

\begin{abstract}
Recently, convolutional neural networks (CNNs) have become popular for image classification due to their effectiveness in computer vision tasks. Now, researchers are exploring the potential of vision transformers (ViTs) in remote sensing and Earth observation. However, traditional Real-Valued networks often overlook important phase information in Complex-Valued (CV) data like polarimetric synthetic aperture radar (PolSAR) data. To address this, new CV deep architectures have emerged. HybridCVNet, a novel hybrid network, blends CV-CNN and CV vision transformer (CV-ViT) techniques. It efficiently combines CV 3D and 2D CNNs as feature extractors, enhancing PolSAR image classification by extracting complementary information and effectively leveraging interdependencies within the data. Experimental results from widely-used PolSAR datasets show HybridCVNet outperforms other methods, achieving an overall accuracy of 97.39\% on the Flevoland dataset and showing promise even with just a 1\% sampling ratio, with a Kappa value of 0.972 on the San Francisco dataset. Source code is accessible through \url{https://github.com/mqalkhatib/HybridCVNet}
\end{abstract}


\section{Introduction} \label{sec:intro}
\vspace{-.5em}
\IEEEPARstart{P}{olarimetric} Synthetic Aperture Radar (PolSAR) images offer a unique perspective in microwave remote sensing, capturing polarization properties for detailed insights into Earth's features like vegetation \cite{hajnsek2009potential} and water bodies \cite{10281440}. Unlike optical images, PolSAR works reliably in all weather conditions and penetrates surfaces effectively. 
Therefore, it is widely employed in both civilian and military applications \cite{lupidi2020polarimetric}.

In recent years, there has been a notable increase in methods proposed for PolSAR image classification. Advancements in machine learning have led to the extensive utilization of various techniques such as K-nearest neighbor (KNN) \cite{ferreira2024machine}, support vector machines (SVMs) \cite{masurkar2024performance}, among others, for this task. Additionally, deep learning models like convolutional neural networks (CNN) \cite{zhou2016polarimetric} gained popularity in PolSAR image classification, showing significant improvements in performance when compared to traditional machine learning frameworks.

In literature, there has been extensive exploration of CNN-based methodologies. The work by \cite{zhou2016polarimetric} initially introduced a CNN model employing real-valued features for PolSAR classification. Subsequently, in \cite{zhang2018polarimetric},A 3D CNN approach was introduced, enabling simultaneous extraction of polarimetric and spatial features. However, these methods solely used real-valued data, overlooking the inherent phase information in PolSAR data. Research has shown that incorporating phase information can notably boost classification effectiveness \cite{zhang2017complex}. Integrating complex-valued (CV) operations within the CNN architecture enhanced performance compared to real-valued networks. Additionally, performance was further boosted by employing CV 3D CNN architectures \cite{tan2019complex}.

CNNs dominate PolSAR classification, but their local processes don't always yield optimal results. Convolution primarily captures local features, limiting global information. Recently, transformers, advanced language models, have shown effectiveness in capturing global correlations in images \cite{dosovitskiy2020image}. Following their success, researchers are now examining the capabilities of transformer models in Earth observation and remote sensing tasks \cite{dong2021exploring}, \cite{10153685}. Vision Transformers (ViTs) were explored by \cite{dong2021exploring}  for the first time for PolSAR classification tasks, results showed that ViT-based methods are better for PolSAR image classification. However, The key issue with transformers is their need for a larger training dataset compared to CNNs. As such, Researchers in \cite{jamali2023local}  proposed an efficient Vision Transformer that incorporates local window attention to accurately classify PolSAR imagery. 

The majority of ViT based methods for PolSAR classification predominantly utilize real-valued representations \cite{dong2021exploring}, \cite{jamali2023local}, thereby overlooking the phase information. Motivated by this, a novel model on using CV-CNN and CV-ViT called HybridCVNet is proposed for PolSAR classification. Unlike existing methods, this approach handles sequences derived from the coherency matrix elements, exploiting correlations between scattering characteristics effectively. Furthermore, by integrating CV layers into ViT, complementary information from different feature maps is extracted, enhancing classification through the capture of discriminative features across various levels. 

The main contributions of this letter are as follows:
\begin{enumerate}
    \item A deep learning-based PolSAR image classification framework was developed, effectively combining both CV-CNNs and CV-ViTs for accurate classification of PolSAR imagery.
    \item To fully leverage the characteristics of PolSAR data, complex-valued (CV) operations are introduced into the ViT architecture to exploit phase information comprehensively. This enhancement further improves the results of PolSAR image classification.
    \item The proposed method is validated using two widely employed PolSAR datasets. The results demonstrate that the method exhibits superior visual performance compared to state-of-the-art  real-valued ViTs, including Swin Transformer and PolSARFormer.
\end{enumerate}

This letter is structured as follows: Section \ref{sec:methodology} presents the HybridCVNet model, Section \ref{sec:results} provides a comprehensive presentation of the results and analyses, and the study concludes in Section \ref{sec:conclusion} with a summary and conclusion remarks.

\section{Methodology} \label{sec:methodology}

\subsection{Polarimetric Data of PolSAR Image}
The properties of how ground objects scatter electromagnetic waves can be explained using polarized scattering matrix $S$ as defined below:
\begin{equation}
S = 
\begin{bmatrix}
S_{HH} & S_{HV}  \\
S_{VH} & S_{VV}  
\end{bmatrix}
,
\end{equation}

where $S_{AB}(A,B \in {H,V})$ represents the backscattering coefficient of the polarized electromagnetic wave in emitting $A$ direction and receiving $B$ direction. $H$ and $V$ represent the horizontal and vertical polarization, respectively. 

\begin{figure*}[!t]
\centering
\includegraphics[width=.90\linewidth]{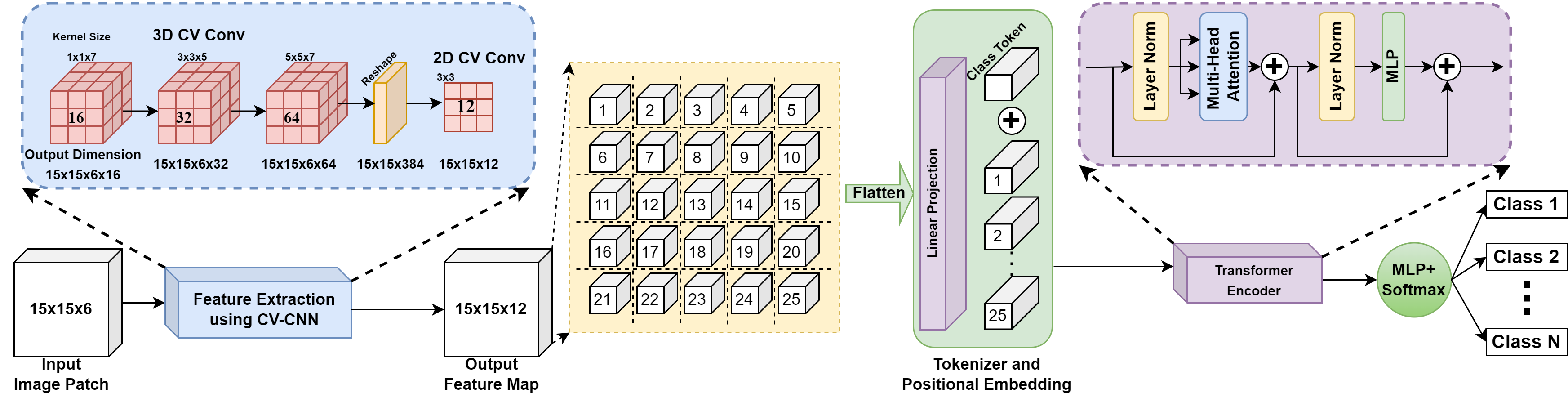}
\vspace{-1em}
\caption{Overall Architecture of HybridCVNet}
 \label{fig:model}
\end{figure*}

Typically, the scattering matrix undergoes a conversion into either the polarization coherency matrix or the polarization covariance matrix. The coherency matrix serves as the most commonly used representation for PolSAR data, as given in Equation \ref{eq:Hermitian matrix} \cite{ren2023new}.
\begin{equation}\label{eq:Hermitian matrix}
    T = \frac{1}{n}\sum_{j = 1}^n \vec{k_j}{\vec{k_j}}^H= \begin{bmatrix}
T_{11} & T_{12}&T_{13}  \\
T_{21} & T_{22}&T_{23}\\
T_{31} & T_{32}&T_{33}
\end{bmatrix},
\end{equation}
\noindent where the operator $(.)^H$ stands for complex conjugate operation
and $n$ is the number of looks. It is worth mentioning that $T$ is a Hermitian matrix with real-valued elements on the diagonal and complex-valued elements off-diagonal. As a result, the three real-valued and three complex-valued elements of the upper triangle of the coherency matrix (i.e. $T11$, $T12$, $T13$, $T22$, $T23$, $T33$) are used as the input features of the models, where each pixel is depicted by a 6-dimensional CV vector. Following that, the PolSAR image is divided into patches using sliding windows for further processing.

\subsection{Feature Extraction using CV-CNN}
CNNs have excelled as feature extractors in various computer vision tasks. However, PolSAR image data, with its complex values, presents challenges for traditional real-valued CNNs. In PolSAR classification, CV-CNNs have shown superiority over traditional CNNs \cite{alkhatib2023polsar}. To leverage CV-CNNs' feature extraction capabilities, a hierarchical architecture inspired by \cite{roy2019hybridsn} was implemented, with all layers converted to complex-valued. The 3D CV-CNN extracts polarimetric-spatial features, while the 2D CV-CNN refines spatial features. This approach ensures computational efficiency, using three 3D convolutional layers (16, 32, 64 kernels) followed by a 2D CV-CNN layer with 12 kernels of size $3 \times 3$.
\highlighting{For an image $X = \Re{(X)} + i\Im{(X)}$ and a kernel $k = \Re{(K)} + i\Im{(K)}$, the result of the complex convolution $Y$ can be expressed in Equation }\ref{eq:complexconv}.

\begin{equation}
  \begin{aligned}
   \label{eq:complexconv}
     Y = X \ast K &= \Re{(X)} \ast \Re{(K)} - \Im{(X)} \ast \Im{(K)}\\ 
     &   + i.(\Re{(X)} \ast \Im{(K)}) + i.(\Im{(X)} \ast \Re{(K)}) 
  \end{aligned}
\end{equation}

 \noindent \highlighting{where $\Re{(.)}$ and $\Im{(.)}$ represent the real and the imaginary parts of a CV number, respectively, $i$ is the imaginary number $\sqrt{-1}$, and $Y$ can be expressed as $\Re{(Y)} + i.\Im{(Y)}$.}

\vspace{-1em}

\subsection{Complex-Valued ViTs}
Vision Transformers (ViTs) have demonstrated remarkable success in computer vision tasks \cite{dosovitskiy2020image} and it showed remarkable results when compared to state-of-the-art CNN models. It is notable also that ViTs are being effectively used in wide range of remote sensing applications, such as classification tasks \cite{roy2023multimodal}. However, the main drawback of transformers is their need for more training data compared to CNNs. Moreover, adapting ViT to handle complex-valued data is crucial for improving transformer performance in PolSAR image classification tasks. To achieve this task, all layers of ViT model were translated to complex layers using CVNN library developed in \cite{barrachina2021complex}, The model will be designated as CV-ViT, standing for \textbf{C}omplex-\textbf{V}alued \textbf{Vi}sion \textbf{T}ransformer. 

\subsection{HybridCVNet}
CV-CNNs excel in capturing local spatial-polarimetric patterns and semantic characteristics, while CV-ViTs specialize in modeling long-range dependencies and global context in images through the self-attention mechanism \cite{dong2021exploring}. By combining the local feature extraction capability of CV-CNNs with the polarimetric global relationship modeling expertise of ViTs, the hybrid model achieves superior performance in PolSAR image classification tasks.

Figure \ref{fig:model} shows the overall architcture of the proposed model. The CV PolSAR image patch, sized $15\times15\times6$ undergoes feature extraction while maintaining its original dimensions. Subsequently, the output from the feature extractor is partitioned into patches of size $3\times3\times6$, resulting in a total of 25 patches. Each patch is then flattened and inputted into the CV-ViT for further processing and classification.

\begin{figure*}[!t]
\centering
\includegraphics[width=.85\linewidth]{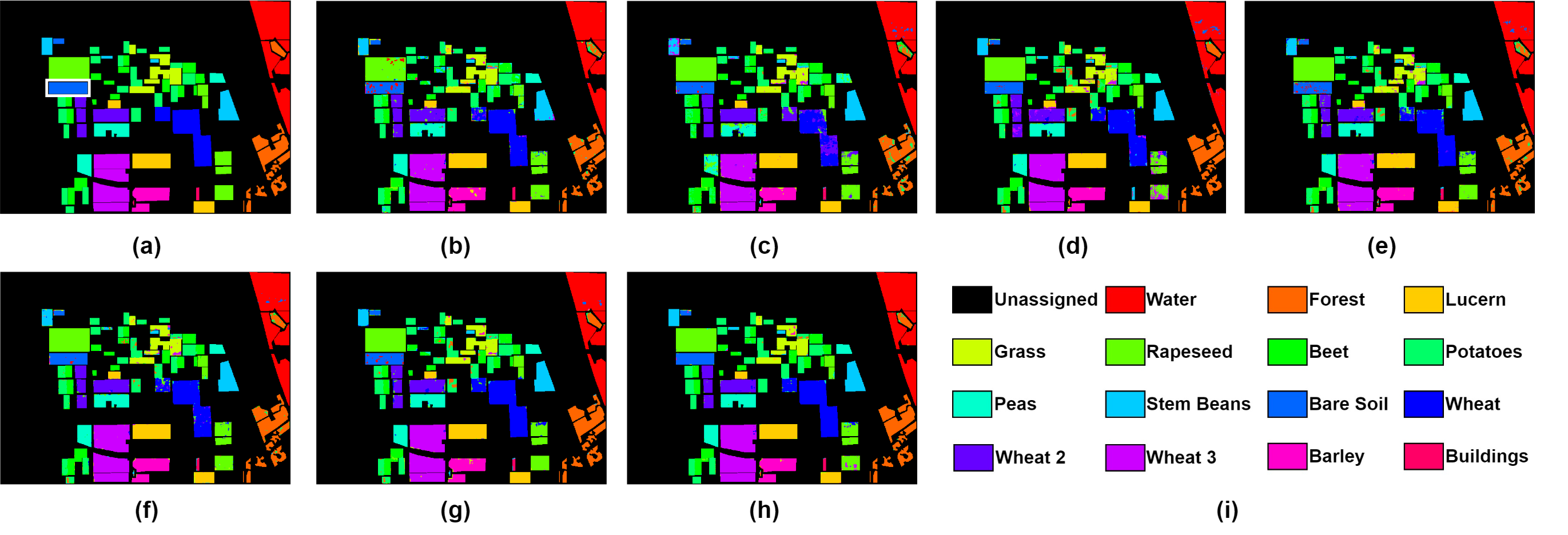}
 \vspace{-1em}
 \caption{Classification results of the Flevoland dataset. (a) Reference Class Map; (b) 3D-CNN; (c) WaveletCNN; (d) ViT; (e) swin Transformer; (f) PolSARFormer; (g) HybridRVNet; (h) HybridCVNet; (i) Classes}
 \label{fig:FL_Results}
\end{figure*}

\begin{table}[t!]
\centering
\caption{Experimental Results of different methods on Felvoland Dataset.}
\label{tab:FL_Results}
\resizebox{\linewidth}{!} {
\begin{tabular}{cccccccc}
\hline
Class              & 3D-CNN         & WaveletCNN     & ViT            & swin           & PolSARFormer   & HybridRVNet    & HybridCVNet             \\ \hline
Water              & 99.46          & 98.59          & 96.93          & 98.23          & 98.23          & 98.99          & \textbf{99.84}          \\
Forest             & 94.56          & 86.87          & \textbf{98.90} & 93.51          & 93.51          & 98.72          & 98.66                   \\
Lucerne            & 96.15          & \textbf{98.50} & 97.61          & 93.42          & 93.42          & 97.04          & 97.46                   \\
Grass              & 85.97          & 89.09          & 84.19          & 79.52          & 79.52          & 88.95          & \textbf{91.47}          \\
Rapeseed           & 94.14          & 95.48          & 88.80          & 96.58          & 96.58          & \textbf{97.12} & 95.15                   \\
Beet               & 81.95          & \textbf{95.74} & 85.81          & 88.45          & 88.45          & 84.91          & 92.57                   \\
Potatoes           & 92.69          & 71.16          & 89.46          & 89.79          & 89.79          & 91.53          & \textbf{97.44}          \\
Peas               & 96.28          & 70.80          & 95.75          & 96.93          & 96.93          & 98.57          & \textbf{99.39}          \\
Stem Beans         & 92.4           & 89.39          & 97.51          & 93.84          & 93.84          & 95.05          & \textbf{98.67}          \\
Bare Soil          & 82.60          & 95.20          & 95.17          & 87.6 0         & 87.6 0         & 91.91          & \textbf{96.18}          \\
Wheat              & 92.45          & 57.96          & 93.56          & 93.42          & 93.42          & 97.94          & \textbf{98.36}          \\
Wheat 2            & 95.29          & \textbf{99.76} & 87.00          & 92.01          & 92.01          & 92.07          & 97.51                   \\
Wheat 3            & 98.66          & 98.16          & 99.38          & 98.01          & 98.01          & 99.24          & \textbf{99.58}          \\
Barley             & 95.55          & 96.66          & 98.77          & 99.13          & \textbf{99.15} & 95.63          & 97.84                   \\
Buildings          & \textbf{98.62} & 74.57          & 9.00           & 65.57          & 65.57          & 69.72          & 86.16                   \\ \hline
OA (\%)            & 93.75$\pm3.23$ & 88.12$\pm2.45$ & 93.28$\pm4.64$ & 93.58$\pm3.87$ & 95.74$\pm2.67$ & 95.39$\pm2.71$ & $\mathbf{97.39\pm1.35}$ \\
AA (\%)            & 93.12$\pm4.01$ & 87.86$\pm3.01$ & 87.85$\pm6.89$ & 91.07$\pm4.13$ & 93.61$\pm3.85$ & 93.15$\pm3.98$ & $\mathbf{96.41\pm2.49}$ \\
Kappa $\times$ 100 & 93.17$\pm2.64$ & 87.01$\pm2.85$ & 92.65$\pm3.01$ & 92.98$\pm3.92$ & 95.35$\pm2.58$ & 94.96$\pm2.92$ & $\mathbf{97.15\pm1.51}$ \\ \hline
\end{tabular}
}
\end{table}

\section{Experimental Results} \label{sec:results}
\vspace{-.5em}
\subsection{Datasets and Settings}
The proposed HybridCVNet model is tested and evaluated on two widely used PolSAR datasets.

The first dataset is the Flevoland dataset, it has a dimension of 750 $\times$ 1024 pixels with 12 meters spatial resolution. It was acquired by the NASA/JPL AIRSAR system on August 16, 1989 for Flevoland area in the Netherland. The dataset contains 15 distinct classes representing various agricultural regions \cite{cao2021polsar}. The Pauli pseudo-color image and ground truth map are shown in Fig. \ref{fig:FL_Results}(a) and (b) respectively.

\begin{figure*}[!t]
\centering
\includegraphics[width=.85\linewidth]{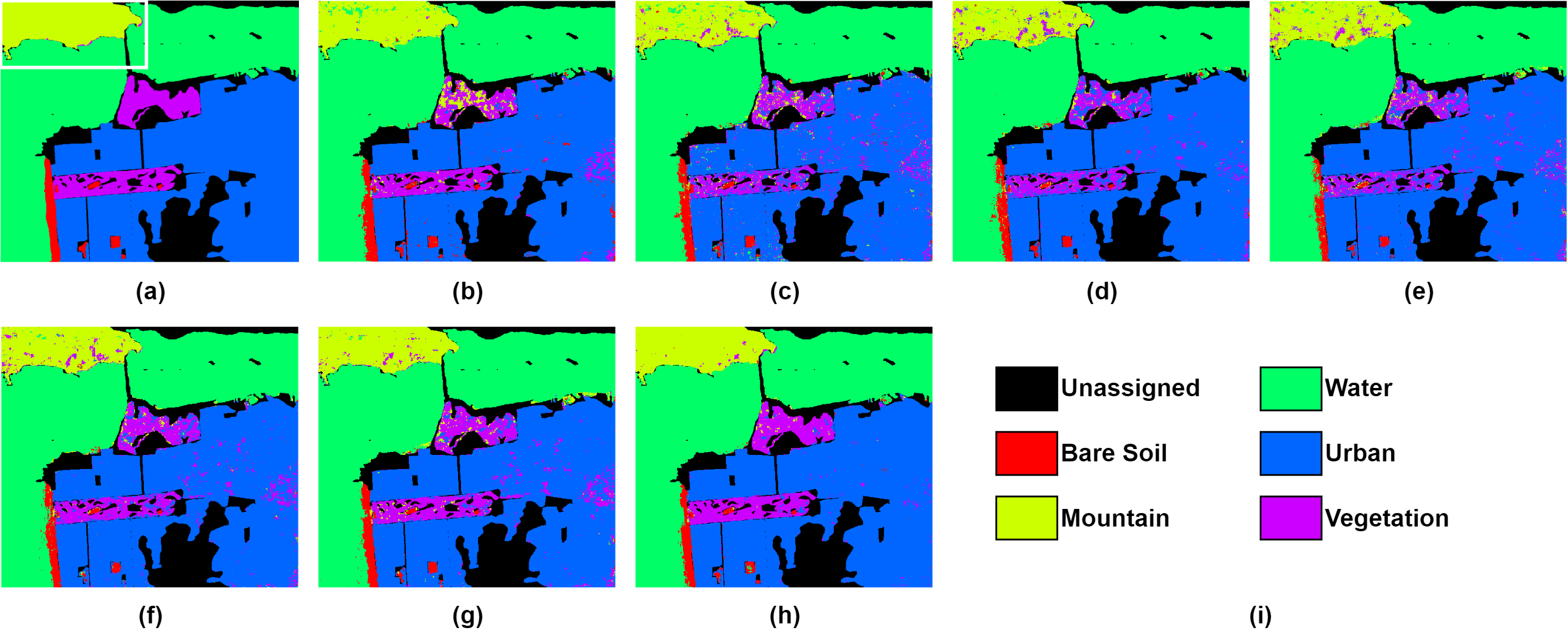}
 \vspace{-1em}
 \caption{Classification results of the San Francisco dataset. (a) Reference Class Map; (b) 3D-CNN; (c) WaveletCNN; (d) ViT; (e) swin Transformer; (f) PolSARFormer; (g) HybridRVNet; (h) HybridCVNet; (i) Classes}
 \label{fig:SF_Results}
\end{figure*}

The second dataset is San Francisco dataset, The image size is $900 \times 1024$ pixels and has a spatial resolution of 10 meters. It was obtained from the L-band AIRSAR over the San Francisco area in 1989. It comprises of five categorized terrain classes \cite{liu2022polsf}. The Pauli pseudo-color image and ground truth map are shown in Fig. \ref{fig:SF_Results}(a) and (b) respectively.

For both datasets, patches of size $15 \times 15$ were extracted. Only 1\% of the data were used for training and the remaining 99\% were used for testing. learning rate is set to $1 \times 10^{-3}$, the batch size is 64, and the training epoch is set to 100. To prevent overfitting, an early stopping strategy was employed. If the model's performance didn't improve over 10 consecutive epochs, training halts, and the model reverts to its optimal weights.

\subsection{Classification Results}
In this section, the Flevoland and San Francisco datasets previously discussed are used to evaluate the classification performance of the proposed model both quantitatively and qualitatively. To ensure consistency in classification outcomes and minimize the impact of random sample selection, the experiments were repeated 10 times. The final result was determined by averaging the outcomes of these experiments, were recorded values are represented in terms of mean and variance. Additionally, detailed classification results for each category were provided.

\begin{figure}[!t]
\centering
\includegraphics[width=.85\linewidth]{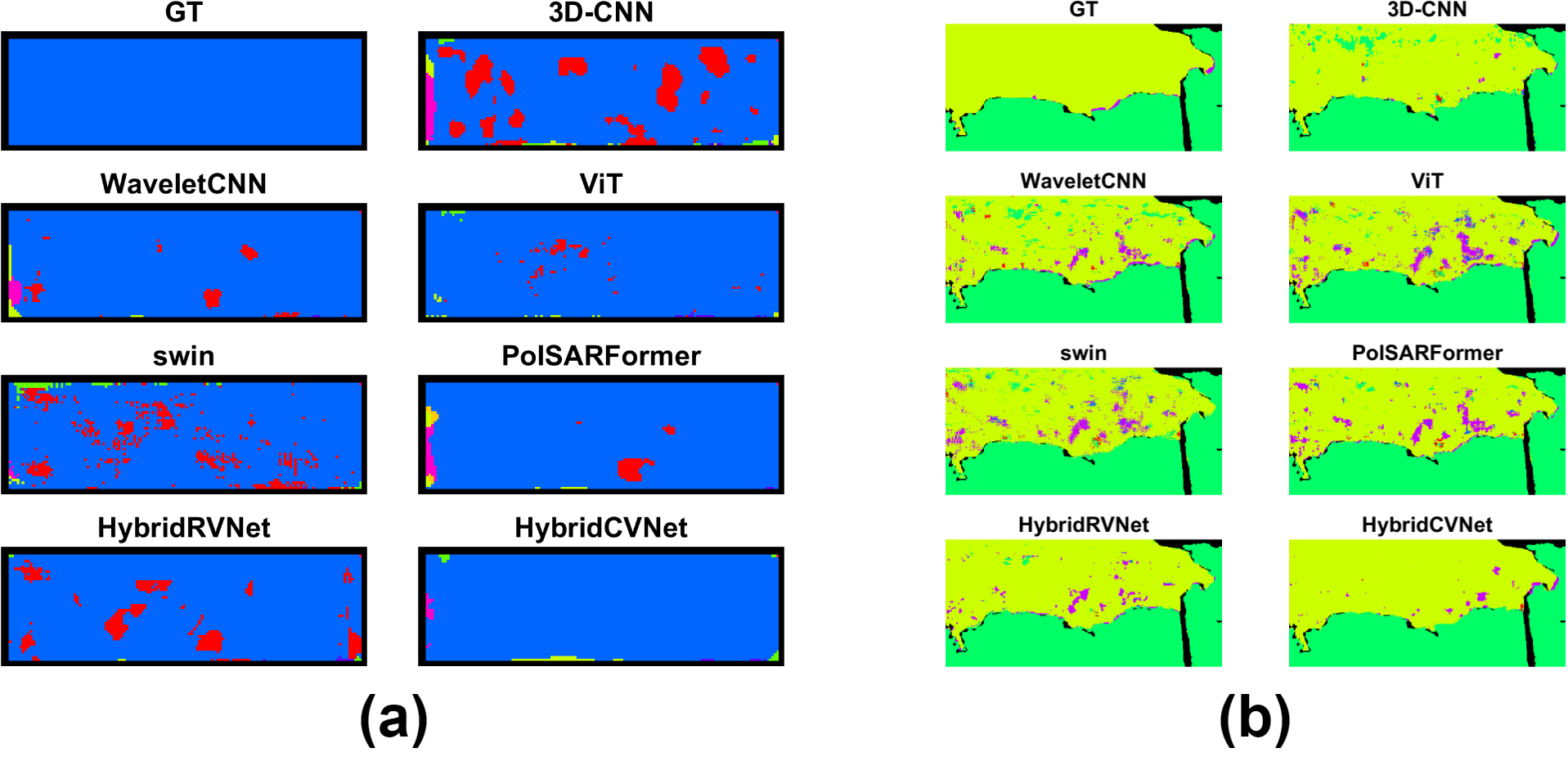}
\vspace{-1em}
\caption{Zoomed area of the white box region in Flevoland region Fig. \ref{fig:FL_Results}(b) and San Francisco region Fig. \ref{fig:SF_Results}(b)   }
 \label{fig:tiles}
\end{figure}

To validate the proposed HybridCVNet, comparison algorithms including 3D Complex-Valued CNN (3D-CNN) \cite{tan2019complex}, \highlighting{WaveletCNN} \cite{jamali2022polsar}, Vision Transformer (ViT) \cite{dosovitskiy2020image}, Swin Transformer (swin) \cite{liu2021swin}, and Local Window Attention Transformer (PolSARFormer) \cite{jamali2023local} are employed. Additionally, a variation of the proposed network involving only real-valued layers (HybridRVNet) is utilized to illustrate the importance of using Complex-Valued layers for PolSAR classification tasks. Table \ref{tab:FL_Results} displays the assessment outcomes on the Flevoland dataset. The results indicate that the HybridCVNet model outperforms other algorithms, including 3D-CNN (93.75\%), WaveletCNN (88.12\%), ViT (93.28\%), swin (93.58\%), PolSARFormer (95.74\%), and HybridRVNet (95.39\%), achieving OA of 97.39\%. The suggested model surpasses the Swin Transformer and PolSARFomer in terms of overall accuracy (OA) by 3.81\% and 1.65\%, respectively. Figure \ref{fig:FL_Results} illustrates the classification maps obtained using different methods for the Flevoland region. As depicted in \ref{fig:tiles}(a), the HybridCVNet network generates a more uniform land cover map with reduced noise compared to the other models. 

\begin{table}[t!]
\centering
\caption{Experimental Results of different methods on San Francisco Dataset.}
\label{tab:SF_Results}
\resizebox{\linewidth}{!} {
\begin{tabular}{cccccccc}
\hline
Class              & 3D-CNN         & WaveletCNN     & ViT            & swin           & PolSARFormer   & HybridRVNet    & HybridCVNet             \\ \hline
Bare Soil          & 85.50          & \textbf{86.74} & 79.96          & 68.16          & 75.29          & 86.33          & 78.80                   \\
Mountain           & 92.48          & 89.45          & 88.77          & 88.85          & 90.93          & 94.10          & \textbf{97.72}          \\
Water              & 99.18          & 99.30          & 99.06          & 99.02          & 99.33          & 98.76          & \textbf{99.54}          \\
Urban              & 96.56          & 92.68          & 97.40          & 95.80          & 96.25          & 95.63          & \textbf{98.82}          \\
Vegetation         & 75.71          & 67.00          & 64.66          & 60.76          & 76.88          & 81.27          & \textbf{91.76}          \\ \hline
OA (\%)            & 95.74$\pm2.92$ & 93.33$\pm2.87$ & 94.92$\pm3.01$ & 93.76$\pm3.25$ & 95.44$\pm3.98$ & 95.68$\pm3.19$ & $\mathbf{98.21\pm0.76}$ \\
AA (\%)            & 89.89$\pm3.73$ & 87.03$\pm3.04$ & 85.97$\pm4.89$ & 82.51$\pm5.03$ & 87.73$\pm5.78$ & 91.21$\pm5.01$ & $\mathbf{93.32\pm3.45}$ \\
Kappa $\times$ 100 & 93.34$\pm2.41$ & 89.61$\pm2.12$ & 91.99$\pm2.91$ & 90.18$\pm4.22$ & 92.86$\pm3.88$ & 93.27$\pm4.04$ & $\mathbf{97.20\pm2.17}$ \\ \hline
\end{tabular}
}
\end{table}

Table \ref{tab:SF_Results} presents the classification outcomes, while Fig. \ref{fig:SF_Results} illustrates the visual classification maps for the San Francisco dataset. The results in Table \ref{tab:SF_Results} indicate that the proposed HybridCVNet model achieves an average accuracy (AA) of 93.32\%, showcasing superior classification outcomes compared to 3D-CNN (89.89\%), WaveletCNN (87.03\%), ViT (85.97\%), swin (82.51\%), PolSARFormer (87.73\%), and HybridRVNet (91.21\%). Although the Swin Transformer algorithm is acknowledged as state-of-the-art ViT, it exhibits lower classification accuracy due to its requirement for a significantly larger amount of training data compared to CNN classifiers. PolSARFomer demonstrates better performance than ViT and swin as it employs CNN as a feature extractor before applying transformer-based classification. However, the proposed HybridCVNet classifier demonstrates notably better PolSAR imagery classification accuracy when compared with the state-of-the-art PolSARFormer. Specifically, HybridCVNet outperforms Swin Transformer and PolSARFormer by approximately 11\% and 6\%, respectively, in terms of AA. As seen in Fig. \ref{fig:tiles}(b), The area enclosed within the white box (Mountain Class) represents the portion that is easily distinguishable, consistent with the findings presented in Table \ref{tab:SF_Results}

\subsection{Ablation study}
To gain deeper insights into the importance of each component of the HybridCVNet model, an ablation study was conducted. The outcomes pertaining to the Flevoland (\textbf{Fl}) and San Francisco (\textbf{SF}) datasets are presented in Table \ref{tab:ablation} using only 1\% of the data for training. Table \ref{tab:ablation} demonstrates that leveraging the capabilities of both CV-CNNs and CV-ViTs resulted in the highest classification accuracy, with an overall accuracy (OA) of 97.15\% on Flevoland, while CV-CNNs and CV-ViTs individually achieved 96.65\% and 95.75\%, respectively. Furthermore, the results from San Francisco indicate that integrating CV-CNNs and CV-ViTs enhanced the accuracy of PolSAR classification by 0.48\% and 1.53\%, respectively, compared to CV-CNNs and CV-ViTs used separately.

\begin{table}[t!]
\centering
\caption{RESULTS OF ABLATION STUDIES ON DIFFERENT COMBINATION OF MODEL components}
\label{tab:ablation}
\resizebox{8.5cm}{!} {
\begin{tabular}{ccccc}
\cline{2-5}
                               & Class          & OA (\%)                 & AA (\%)                 & Kappa $\times$ 100      \\ \hline
\multirow{3}{*}{\rotatebox{90}{\textbf{Fl} }}   & CV-CNN         & 96.93 $\pm$1.44         & 95.55$\pm$2.50          & 96.65$\pm$1.55          \\
                               & CV-ViT & 96.11$\pm$1.69          & 94.47$\pm$2.45          & 95.75$\pm$1.75          \\
                               & HybridCVNet    & $\mathbf{97.39\pm1.35}$ & $\mathbf{96.41\pm2.49}$ & $\mathbf{97.15\pm1.51}$ \\ \hline
\multirow{3}{*}{\rotatebox{90}{\textbf{SF} }} & CV-CNN         & 97.92$\pm$1.08          & 91.94$\pm$3.55          & 96.72$\pm$2.32          \\
                               & CV-ViT & 97.25$\pm$1.30          & 89.06$\pm$3.68          & 95.67$\pm$2.47          \\
                               & HybridCVNet    & $\mathbf{98.21\pm0.76}$ & $\mathbf{93.32\pm3.45}$ & $\mathbf{97.20\pm2.17}$ \\ \hline
\end{tabular}
}
\end{table}

Additionally, Table \ref{tab:ablation} illustrates that CV-CNNs surpass CV-ViTs in terms of accuracy. Transformer-based classifiers are recognized for their requirement of significant amount of data during training.

\subsection{Models Performance at Different Percentages of Training Data}
\begin{figure*}[!t]
\centering
\includegraphics[width=.95\linewidth]{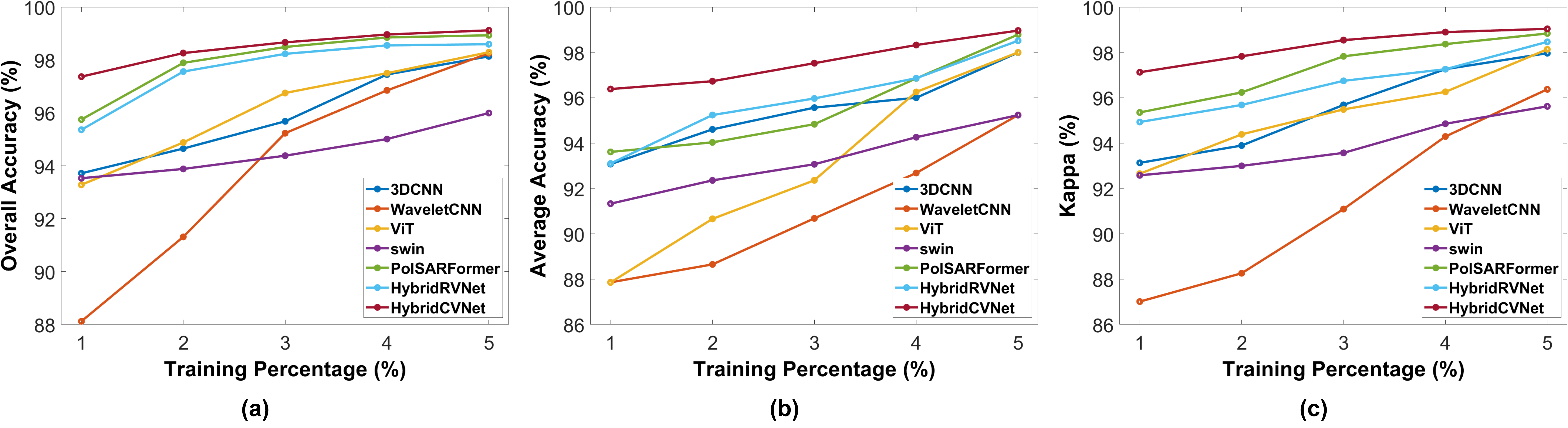}
\vspace{-1em}
\caption{Classification accuracy of Flevoland dataset at different percentages of training data (a) OA (b) AA and (c) Kappa index.}
 \label{fig:FL_Percentages}
\end{figure*}

The assessment of the model's performance is effectively achieved by examining its classification accuracy with varying percentages of training data. Random subsets comprising 1\%, 2\%, 3\%, 4\%, and 5\% of labeled samples were selected for training, while the remaining samples were utilized to evaluate the model's performance. The classification outcomes (Overall Accuracy, Average Accuracy and Kappa index) are illustrated in Fig. \ref{fig:FL_Percentages}. It's observable that, the classification accuracy improves as the number of training samples increases. It's evident that when the training percentage is low, CNN-based models outperform Transformer-based classification models like ViT and Swin, mainly because the latter demand a significant amount of training data compared to CNN classifiers. However, as the number of training samples increases, ViT starts to surpass 3D-CNN in terms of overall accuracy (OA), average accuracy (AA), and Kappa. Moreover, the findings indicate that the classification accuracy of the HybridCVNet algorithm in terms of OA, AA, Kappa increased by around 1.75\%, 2.58\%, and 1.91\%, respectively, when utilizing a training data ratio of 1\%–5\%, as depicted in Figure \ref{fig:FL_Percentages}.

\section{Conclusion} \label{sec:conclusion}
\vspace{-.5em}
This letter presents HybridCVNet, a novel approach for PolSAR image classification that combines CV-CNN and CV-ViT architectures to enhance feature extraction capabilities. Evaluations on two benchmark datasets demonstrate HybridCVNet's superiority over current state-of-the-art models. On the Flevoland dataset, HybridCVNet achieves OA improvements of 2.8\% and 5.34\% over PolSARFormer and Swin Transformer, respectively. Additionally, on the San Francisco dataset, HybridCVNet outperforms ViT, Swin Transformer, and PolSARFormer, attaining AA of 93.32\%, compared to PolSARFormer's 87.73\%.

\highlighting{Although the model outperforms state-of-the-art approaches, it has higher computational costs and longer training times due to complex-valued operations. This was partly mitigated with early stopping. Future work will explore model pruning and knowledge distillation to reduce complexity and inference time while preserving accuracy.
}
\bibliographystyle{IEEEtran}
\balance
\bibliography{bibtex/Main_Doc}

@article{cao2021polsar,
  title={PolSAR image classification using a superpixel-based composite kernel and elastic net},
  author={Cao, Yice and Wu, Yan and Li, Ming and Liang, Wenkai and Zhang, Peng},
  journal={Remote Sensing},
  volume={13},
  number={3},
  pages={380},
  year={2021},
  publisher={MDPI}
}

@article{jamali2022polsar,
  title={PolSAR image classification based on deep convolutional neural networks using wavelet transformation},
  author={Jamali, Ali and Mahdianpari, Masoud and Mohammadimanesh, Fariba and Bhattacharya, Avik and Homayouni, Saeid},
  journal={IEEE Geoscience and Remote Sensing Letters},
  volume={19},
  pages={1--5},
  year={2022},
  publisher={IEEE}
}

@ARTICLE{10153685,
  author={Roy, Swalpa Kumar and Deria, Ankur and Hong, Danfeng and Rasti, Behnood and Plaza, Antonio and Chanussot, Jocelyn},
  journal={IEEE Transactions on Geoscience and Remote Sensing}, 
  title={Multimodal Fusion Transformer for Remote Sensing Image Classification}, 
  year={2023},
  volume={61},
  number={},
  pages={1-20},
  doi={10.1109/TGRS.2023.3286826}}

@inproceedings{liu2022polsf,
  title={PolSF: PolSAR image datasets on san Francisco},
  author={Liu, Xu and Jiao, Licheng and Liu, Fang and Zhang, Dan and Tang, Xu},
  booktitle={International Conference on Intelligence Science},
  pages={214--219},
  year={2022},
  organization={Springer}
}

@article{tan2019complex,
  title={Complex-valued 3-D convolutional neural network for PolSAR image classification},
  author={Tan, Xiaofeng and Li, Ming and Zhang, Peng and Wu, Yan and Song, Wanying},
  journal={IEEE Geoscience and Remote Sensing Letters},
  volume={17},
  number={6},
  pages={1022--1026},
  year={2019},
  publisher={IEEE}
}

@inproceedings{liu2021swin,
  title={Swin transformer: Hierarchical vision transformer using shifted windows},
  author={Liu, Ze and Lin, Yutong and Cao, Yue and Hu, Han and Wei, Yixuan and Zhang, Zheng and Lin, Stephen and Guo, Baining},
  booktitle={Proceedings of the IEEE/CVF international conference on computer vision},
  pages={10012--10022},
  year={2021}
}

@article{dosovitskiy2020image,
  title={An image is worth 16x16 words: Transformers for image recognition at scale},
  author={Dosovitskiy, Alexey and Beyer, Lucas and Kolesnikov, Alexander and Weissenborn, Dirk and Zhai, Xiaohua and Unterthiner, Thomas and Dehghani, Mostafa and Minderer, Matthias and Heigold, Georg and Gelly, Sylvain and others},
  journal={arXiv preprint arXiv:2010.11929},
  year={2020}
}

@article{jamali2023local,
  title={Local window attention transformer for polarimetric SAR image classification},
  author={Jamali, Ali and Roy, Swalpa Kumar and Bhattacharya, Avik and Ghamisi, Pedram},
  journal={IEEE Geoscience and Remote Sensing Letters},
  volume={20},
  pages={1--5},
  year={2023},
  publisher={IEEE}
}

@article{ren2023new,
  title={A new architecture of a complex-valued convolutional neural network for polsar image classification},
  author={Ren, Yihui and Jiang, Wen and Liu, Ying},
  journal={Remote Sensing},
  volume={15},
  number={19},
  pages={4801},
  year={2023},
  publisher={MDPI}
}

@inproceedings{alkhatib2023polsar,
  title={Polsar image classification using attention based shallow to deep convolutional neural network},
  author={Alkhatib, Mohammed Q and Al-Saad, Mina and Aburaed, Nour and Zitouni, M Sami and Al-Ahmad, Hussain},
  booktitle={IGARSS 2023-2023 IEEE International Geoscience and Remote Sensing Symposium},
  pages={8034--8037},
  year={2023},
  organization={IEEE}
}

@article{roy2019hybridsn,
  title={HybridSN: Exploring 3-D--2-D CNN feature hierarchy for hyperspectral image classification},
  author={Roy, Swalpa Kumar and Krishna, Gopal and Dubey, Shiv Ram and Chaudhuri, Bidyut B},
  journal={IEEE Geoscience and Remote Sensing Letters},
  volume={17},
  number={2},
  pages={277--281},
  year={2019},
  publisher={IEEE}
}

@inproceedings{barrachina2021complex,
  title={Complex-valued vs. real-valued neural networks for classification perspectives: An example on non-circular data},
  author={Barrachina, Jose Agustin and Ren, Chenfang and Morisseau, Christele and Vieillard, Gilles and Ovarlez, J-P},
  booktitle={ICASSP 2021-2021 IEEE International Conference on Acoustics, Speech and Signal Processing (ICASSP)},
  pages={2990--2994},
  year={2021},
  organization={IEEE}
}

@article{roy2023multimodal,
  title={Multimodal fusion transformer for remote sensing image classification},
  author={Roy, Swalpa Kumar and Deria, Ankur and Hong, Danfeng and Rasti, Behnood and Plaza, Antonio and Chanussot, Jocelyn},
  journal={IEEE Transactions on Geoscience and Remote Sensing},
  year={2023},
  publisher={IEEE}
}

@article{dong2021exploring,
  title={Exploring vision transformers for polarimetric SAR image classification},
  author={Dong, Hongwei and Zhang, Lamei and Zou, Bin},
  journal={IEEE Transactions on Geoscience and Remote Sensing},
  volume={60},
  pages={1--15},
  year={2021},
  publisher={IEEE}
}

@article{hajnsek2009potential,
  title={Potential of estimating soil moisture under vegetation cover by means of PolSAR},
  author={Hajnsek, Irena and Jagdhuber, Thomas and Schon, Helmut and Papathanassiou, Konstantinos Panagiotis},
  journal={IEEE Transactions on Geoscience and Remote Sensing},
  volume={47},
  number={2},
  pages={442--454},
  year={2009},
  publisher={IEEE}
}

@inproceedings{lupidi2020polarimetric,
  title={Polarimetric radar technology for european defence superiority-the polrad project},
  author={Lupidi, Alberto and Greiff, Christian and Br{\"u}ggenwirth, Stefan and Brandfass, Michael and Martorella, Marco},
  booktitle={2020 21st International Radar Symposium (IRS)},
  pages={6--10},
  year={2020},
  organization={IEEE}
}

@article{ferreira2024machine,
  title={Machine learning classification based on k-Nearest Neighbors for PolSAR data},
  author={Ferreira, Jodavid A and Rodrigues, Anny KG and Ospina, Raydonal and Gomez, Luis},
  journal={Anais da Academia Brasileira de Ci{\^e}ncias},
  volume={96},
  number={1},
  pages={e20230064},
  year={2024},
  publisher={SciELO Brasil}
}

@article{masurkar2024performance,
  title={Performance analysis of SAR filtering techniques using SVM and Wishart Classifier},
  author={Masurkar, Akhil and Daruwala, Rohin and Mohite, Arya},
  journal={Remote Sensing Applications: Society and Environment},
  volume={34},
  pages={101189},
  year={2024},
  publisher={Elsevier}
}

@INPROCEEDINGS{10281440,
  author={Al-Saad, Mina and Aburaed, Nour and Zitouni, M. Sami and Alkhatib, Mohammed Q. and Almansoori, Saeed and Al Ahmad, Hussain},
  booktitle={IGARSS 2023 - 2023 IEEE International Geoscience and Remote Sensing Symposium}, 
  title={A Robust Change Detection Methodology for Flood Events Using SAR Images}, 
  year={2023},
  volume={},
  number={},
  pages={341-344},
  doi={10.1109/IGARSS52108.2023.10281440}}

@article{zhou2016polarimetric,
  title={Polarimetric SAR image classification using deep convolutional neural networks},
  author={Zhou, Yu and Wang, Haipeng and Xu, Feng and Jin, Ya-Qiu},
  journal={IEEE Geoscience and Remote Sensing Letters},
  volume={13},
  number={12},
  pages={1935--1939},
  year={2016},
  publisher={IEEE}
}

@inproceedings{zhang2018polarimetric,
  title={Polarimetric SAR terrain classification using 3D convolutional neural network},
  author={Zhang, Lamei and Chen, Zexi and Zou, Bin and Gao, Ye},
  booktitle={IGARSS 2018-2018 IEEE International Geoscience and Remote Sensing Symposium},
  pages={4551--4554},
  year={2018},
  organization={IEEE}
}

@article{zhang2017complex,
  title={Complex-valued convolutional neural network and its application in polarimetric SAR image classification},
  author={Zhang, Zhimian and Wang, Haipeng and Xu, Feng and Jin, Ya-Qiu},
  journal={IEEE Transactions on Geoscience and Remote Sensing},
  volume={55},
  number={12},
  pages={7177--7188},
  year={2017},
  publisher={IEEE}
}
\end{document}